\documentclass[sigconf]{acmart}

\usepackage{csquotes}
\usepackage{footnote}
\makesavenoteenv{tabular}
\makesavenoteenv{table}
\usepackage{tablefootnote}
\usepackage{lipsum,afterpage,refcount}
\newcommand{\setfootnotemark}{%
  \refstepcounter{footnote}%
  \footnotemark[\value{footnote}]}
  
\def \hfillx {\hspace*{-\textwidth} \hfill}
\AtBeginDocument{%
  \providecommand\BibTeX{{%
    \normalfont B\kern-0.5em{\scshape i\kern-0.25em b}\kern-0.8em\TeX}}}

\copyrightyear{2021}
\acmYear{2021}
\setcopyright{acmcopyright}\acmConference[CODS COMAD 2021]{8th ACM IKDD CODS and 26th COMAD}{January 2--4, 2021}{Bangalore, India}
\acmBooktitle{8th ACM IKDD CODS and 26th COMAD (CODS COMAD 2021), January 2--4, 2021, Bangalore, India}
\acmPrice{15.00}
\acmDOI{10.1145/3430984.3431007}
\acmISBN{978-1-4503-8817-7/21/01}

\begin{document}

\title{Fine-Tune Longformer for Jointly Predicting Rumor Stance and Veracity}

\author{Anant Khandelwal}
\email{anant.iitd.2085@gmail.com}
\affiliation{%
  \institution{Senior Data Scientist, [24]7.ai}
  \city{Bangalore, India}
}

\renewcommand{\shortauthors}{Khandelwal, et al.}

\begin{abstract}
Increased usage of social media caused the popularity of news and events that are not even verified, resulting in the spread of rumors all over the web. Due to widely available social media platforms and increased usage caused the data to be available in large amounts. The manual methods to process such data is costly and time-taking, so there has been increased attention to process and verify such content automatically for the presence of rumors. Lots of research studies reveal that identifying the stances of posts in the discussion thread of such events and news is an important preceding step before detecting the rumor veracity. In this paper, we propose a multi-task learning framework for jointly predicting rumor stance and veracity on the dataset released at SemEval 2019 RumorEval: Determining rumor veracity and support for rumors (SemEval 2019 Task 7), which includes social media rumors stem from a variety of breaking news stories from Reddit as well as Twitter. Our framework consists of two parts: a) The bottom part of our framework classifies the stance for each post in the conversation thread (discussing a rumor) via modeling the multi-turn conversation so that each post aware of its neighboring posts. b) The upper part predicts the rumor veracity of the conversation thread respecting the stance evolution obtained from the bottom part. Experimental results on SemEval 2019 Task 7 dataset show that our method outperforms previous methods on both rumor stance classification and veracity prediction.
\end{abstract}

\keywords{Rumor Stance, Rumor Veracity, Longformer, Roberta, Transformer, Multi-turn Conversation, Model Averaging}

\maketitle

\section{Introduction}
\label{intro}
Social media platforms like Twitter, Reddit, etc generate large amounts of data continuously\cite{hamidian2019rumor, li-etal-2019-rumor-detection, zhang2015automatic}. Nowadays, any novel breaking news appears first on these platforms\cite{kumar2014detecting}. 
Wide usage of social media platforms results in rapid spread and propagation of unverified content i.e.   \textit{rumor} \cite{zubiaga2016analysing, kumar2014detecting, kumar2018false, ma2017detect, popat2017truth, baly2018predicting, zhang2015automatic, zubiaga2016stance, zubiaga2018detection}. The rumor has been defined as \enquote{\textit{a claim whose truthfulness is in doubt and has no clear source, even if its ideological or partisan origins and intents are clear}} \cite{harsin2006rumour}. Rumors bring about harmful effects like spreading fear or even euphoria, cause people to make a wrong judgment, cause damages to political events, economy, and social stability\cite{rubin2017deception}. The massive increase of the social media data rendered the manual methods to debunk the rumors, difficult and costly. Machine Learning and Deep Learning, based methods to identify such phenomena have attracted more attention to the research community in recent years. A Rumor resolution pipeline contains several components such as rumor detection, rumor tracking, and stance classification leading to the final task of determining the veracity of a rumor \cite{zubiaga2018detection}. In this paper, we concentrate on tasks A and B released in SemEval2019 Task 7 (RumorEval 2019) i.e. Stance classification and Veracity prediction respectively\cite{gorrell-etal-2019-semeval}. 

Fine-grained definition for stance classification and veracity prediction is provided by the organizers of RumorEval 2019\cite{gorrell-etal-2019-semeval}. The detailed description for each of the two tasks is described as follows:
\begin{itemize}
    \item \textbf{Sub-Task A}: Given a conversation thread discussing the claim starting from source post, each of the posts in the thread are classified into four labels namely \textbf{S}upport, \textbf{D}eny, \textbf{Q}uery and \textbf{C}omment (SDQC).  
    \item \textbf{Sub-Task B}: Given the source post that started the conversation discussing the rumor is classified as True, False and Unverified.
\end{itemize}
The dataset released for this task\cite{gorrell-etal-2019-semeval} consists of conversations threads from Twitter and Reddit. Reddit threads tend to be longer and more diverse, causing posts in discussion threads to be loosely connected to source post, 
making the task more challenging\cite{gorrell-etal-2019-semeval}.

A lot of the systems submitted to RumorEval 2019\cite{gorrell-etal-2019-semeval} have used an ensemble model for the task. For example, the best performing system in Subtask B (eventAI)\cite{li-etal-2019-eventai}, which implemented an ensemble of classifiers (SVM, RF, LR) with features obtained from LSTM attention network and other range of features characterizing posts and conversation. The second best performing system in sub-task A (BUT-FIT)\cite{fajcik-etal-2019-fit} uses an ensemble of BERT\cite{devlin2018bert} models (with pre-training) for different parameter settings. The best performing system in task A (BLCU-NLP)\cite{yang2019blcu_nlp} and the third best (CLEARumor)\cite{baris2019clearumor} also uses pre-trained representation with OpenAI GPT\cite{radford2018improving} and ELMO\cite{peters2018deep} respectively. Most of the systems use a single post or pair of posts(source \& response) as their input, BLCU-NLP\cite{yang2019blcu_nlp} uses the complete conversation thread starting from the source post with subsequent replies in the temporal order. They also augment the training data with more conversations from external public datasets to increase the generalizability of the model. Most of the above-discussed systems have high performance on either (a) Sub-Task A or (b) Sub-Task B (given in the RumorEval 2019). This may be due to the different text style, level of complexities, and lengths of conversations obtained from Twitter and Reddit.

Apart from the systems submitted to RumorEval 2019, other studies have observed that stance classification towards rumors is viewed as an important preceding step of rumor veracity prediction \cite{wei2019modeling, qazvinian-etal-2011-rumor, 10.1145/2736277.2741637, 10.1145/1964858.1964869, procter2013reading, 10.1145/2806416.2806651, jin2016news, glenski-etal-2018-identifying}, especially in the context of Twitter conversations \cite{zubiaga2016stance}. The approach proposed in\cite{wei2019modeling} is based on the fact that the temporal dynamics of stances indicate rumor veracity. They have observed that there can be conversations that start with a supporting stance ( indicate rumor ) but as the stance evolves with the conversation, the deny stance indicates the false rumor (see Figure. \ref{fig:case}). Based on this observation, we propose to use the post-level (sentence-level) encoder to learn the temporal dynamics of stance evolution for effective veracity prediction (see Section \ref{sentenc}).

Most of the top performing methods\cite{gorrell-etal-2019-semeval}  at RumorEval 2019 perform well on either task i.e. stance classification or veracity prediction separately but not both, which is sub-optimal and limits the generalization of models. But as we have observed previously, these two tasks are inter-related because the stance evolution can provide indicative clues to facilitate veracity prediction. Thus, we proposed a joint learning framework for these two tasks to make better use of their interrelation.

Based on the above context we can divide the problem into following sub-problems a) to leverage the stance evolution for effective veracity prediction b) to identify the stance of the current post, given the neighboring posts in the conversation thread c) to extract the features for each post in conversation thread and also peculiar to a tree-structured conversation, which can provide extra clues for determining stance and veracity like user features (verified user, profile pic present, previous history, etc.), structural features(like retweet count, number of hashtags, number of question marks, URL count, etc) content features (like false synonym, false antonym, etc) and psycho-linguistic features ( emotion feature using Emolex\cite{mohammad2013crowdsourcing, plutchik2001nature}, Emotion Sentiment\cite{poria2013enhanced, ekman1992argument}, post-depth in conversation tree, Speech-act features like order, accept etc). Our approach is based on four main ideas:
\begin{itemize}
    \item \textbf{Multi-turn Conversational Modelling: } The goal is to learn the stance evolution during the conversation to determine the rumor veracity accurately. Also, the stance of a post depends on the neighboring post. So we choose the variant of Transformer\cite{vaswani2017attention} called Longformer\cite{beltagy2020longformer} (trained from   RoBERTa\cite{liu2019roberta} weights) having maximum position embeddings up to 4096 (compared to other pre-trained models like BERT\cite{devlin2018bert} which have 512 and Open-AI GPT-2\cite{radford2019language} which have 1024). This allows us to use Longformer as a base model which additionally has the facility to configure the window of self-attention while modeling the whole conversation at one go. With a sliding window-based attention mechanism it can capture signals from a neighboring post, and due to stacked layers, it has a large receptive field like in the case of stacked CNN’s\cite{wu2019pay}. Additionally we have used, various sentence encoders(like LSTM\cite{doi:10.1162/neco.1997.9.8.1735}, Transformer\cite{vaswani2017attention} etc.) to establish the inter-sentence relation between posts to learn the fine-tuned representation specific to rumor stance and veracity. 
    \item \textbf{Exploiting word-level, post-level and psycho-linguistic features: } The goal is to minimize the direct dependencies on the in-depth grammatical structure of conversations from different social media. We have extracted several stylistic and structural features characterizing Twitter or Reddit language. Also, we utilize conversational-based features to capture the tree-structure of the dataset. We have also used affective and emotion-based feature by extracting information from several resources like LIWC\cite{pennebakerlinguistic}, EmoSenticNet\cite{poria2013enhanced, ekman1992argument}, Emolex\cite{mohammad2013crowdsourcing, plutchik2001nature} and ANEW\cite{osgood1957measurement}. Additionally, speech-act features from\cite{wierzbicka1987english} having 229 verbs across 37 categories. We use the term \enquote{NLP Features} to represent them in the entire paper.
    \item \textbf{Jointly Learning the Rumor Stance and Veracity Classification: } Since the stance evolution across the conversation provide indicative clues to predict the rumor veracity, we have learned the two tasks jointly to establish the inter-relation between them.
    \item \textbf{Ensembling: } We have trained different models by varying the type of sentence encoder and learning rate for each configuration and save them to create a pool of models. To increase the F1-measure and reduce overfitting we have used the Top-$N_s$ fusion strategy used in\cite{fajcik-etal-2019-fit} to select the best models from the pool of saved models.
\end{itemize}
The intuition behind the \enquote{NLP features} are the following:
\begin{itemize}
    \item \textbf{Structural Features}: Like in \cite{kochkina2017turing, pamungkas2019stance, li-etal-2019-eventai}, these are extracted to capture the data characteristics like average word length, ratio of capital words. And, count of chars, words, URLs, hashtags, question marks, periods, etc. And, a boolean feature indicating the presence of exclamation mark, negative words, media content like images video, etc, question mark, and additional to previous work also extracted 37 pos tags\footnote{36 from https://www.ling.upenn.edu/courses/Fall\_2003/ling001/penn\_treebank\_pos.html \& additional ’X’ from https://spacy.io/api/annotation\#pos-tagging}. Further, boolean features include a flag to indicate post is the source post. Additionally, a feature indicating the presence of rumor words like gossip, hoax, etc. or words indicating doubt in the certainty of an assertion like unconfirmed.
     \item \textbf{Content Features: } We have extracted the count of false synonyms, antonyms and question word for each post and concatenated with that of source post every time unlike\cite{yang2019blcu_nlp} which extracted these only for one post at a time. Also, leveraged external resources\cite{bahuleyan-vechtomova2017semeval} to identify presence of cue\cite{bahuleyan-vechtomova2017semeval} and swear words\footnote{https://www.cs.cmu.edu/~biglou/resources/bad-words.txt}.
    \item \textbf{Conversational features: }
    These features depict the tree-structure of a conversation thread. Different from\cite{pamungkas2019stance} we have used embeddings instead of count statistics. We have leveraged embeddings based on Paragram\cite{wieting2015towards} instead of Word2vec as in \cite{kochkina2017turing}. Specifically, we extracted average word embeddings based on paragram, paragram vector similarity with source, prev, and other content of the conversation (obtained by concatenating all other posts in time sequence) and normalized depth of each post in the tree.
    
    \item \textbf{Affective features: } Like in \cite{pamungkas2019stance} we have extracted three features from DAL\cite{whissell2009using} and three feature from affective norm rating (ANEW)\cite{osgood1957measurement}, and an additional AFINN\cite{nielsen2011afinn} sentiment score different from\cite{pamungkas2019stance}.
    \item \textbf{Emotion Features: } These features help to focus on emotional responses to true and false rumors, similar to\cite{pamungkas2019stance} i.e. EmoLex\cite{mohammad2013crowdsourcing, plutchik2001nature} and EmoSenticNet\cite{ekman1992argument, poria2013enhanced}.
    \item \textbf{LIWC(Linguistic Inquiry and Word Count): } We have extracted 2 sentiment category feature (PosEMO \& NegEMO) and 11 categories\cite{pennebakerlinguistic} specific to each stance as described in\cite{pamungkas2019stance}, which extracted these eleven as dialogue-act features.
    \item \textbf{Speech-Act features: } Different from\cite{pamungkas2019stance} which extracted eleven categories from LIWC\cite{pamungkas2019stance}, we have extracted speech-act features specific to verb from\cite{wierzbicka1987english} consisting of 37 categories compiling a total of 229 verbs representing different speech acts. 
\end{itemize}

\begin{figure*}[]
    \centering
    \includegraphics[width=0.8\linewidth]{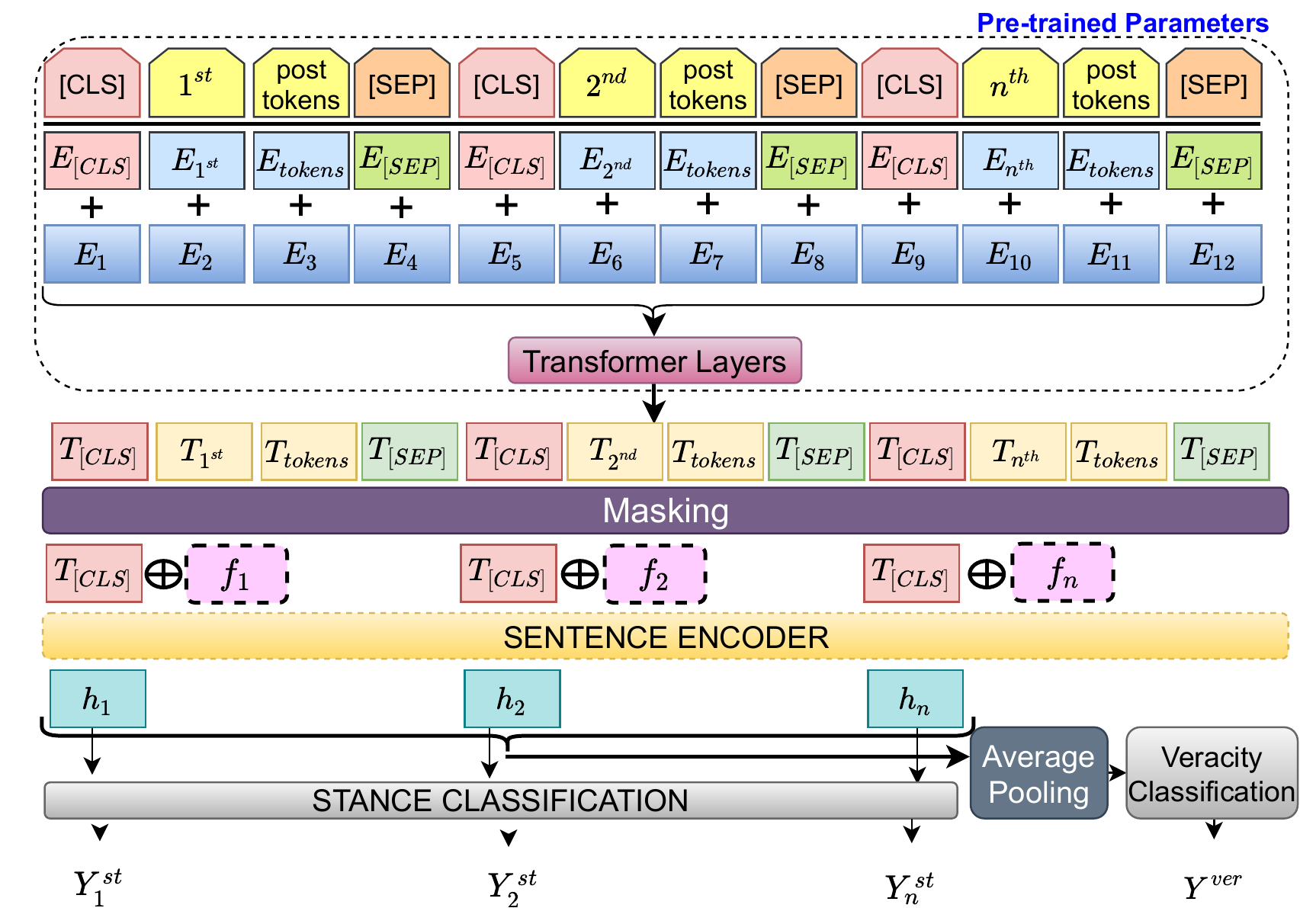}
    \caption{Model Architecture: Each of the individual post is tokenized and the corresponding tokens are separated by [SEP] token. All the post are arranged in the temporal order and the "NLP Features" extracted for each post, are passed through a linear layer denoted as $f_1, f_2,...f_n$. Each of the feature $f_i$ is concatenated with the post representation obtained at the output of Longformer\cite{beltagy2020longformer} at the corresponding [CLS] token. Finally, stance labels are denoted by $Y_i^{st}$ \& veracity label is denoted by $Y^{ver}$}
    \label{fig:arch}
    \vspace{-1.2mm}
\end{figure*}

The model architecture of our proposed system is shown in Figure \ref{fig:arch}. The proposed system does not use any data augmentation techniques like\cite{yang2019blcu_nlp}, which is the top performer in RumorEval 2019 (in Sub-Task A). They also used the features different for Twitter and Reddit. This means the performance achieved by our system solely depends on the training dataset provided by RumorEval 2019. Also, we have used the unified set of features for both Twitter and Reddit. This also proves the effectiveness of our approach. Our system outperforms all the previous state of the art approaches used for rumor stance and veracity prediction on RumorEval 2019 data. Also, our approach outperforms on both tasks compared to other top-performing systems in RumorEval 2019\cite{gorrell-etal-2019-semeval} which performs well on either of the tasks. The remaining part of this paper is organized as follows: Section \ref{relwork} is an overview of related work. Section \ref{probdef} define the problem for each task, Section \ref{propm} describes our proposed method in detail. Section\ref{exp} discusses the experimental evaluation
of the system, and finally, Section \ref{conc} concludes this paper.

\section{Related Work}
\label{relwork}
Within NLP research the task to identify the false content, stance classification of news articles, fact checking etc. has gained momentum to build the automatic system\cite{gorrell-etal-2019-semeval, derczynski-etal-2017-semeval, mohammad-etal-2016-semeval}. Initial work on rumor detection and stance classification\cite{qazvinian-etal-2011-rumor, zhang2015automatic, 10.1145/1964858.1964869, 10.1145/2806416.2806651, procter2013reading} was succeeded by more sophisticated systems\cite{mohammad-etal-2016-semeval, augenstein-etal-2016-stance, mohtarami-etal-2018-automatic, gorrell-etal-2019-semeval, glenski-etal-2018-identifying, derczynski-etal-2017-semeval}.  
Stance analysis has been widely studied in different contexts\cite{somasundaran-wiebe-2009-recognizing, hasan-ng-2013-extra}. Specifically, the studies which classify stance towards rumors in social media\cite{10.1145/1964858.1964869, qazvinian-etal-2011-rumor, procter2013reading}. Some proposed linguistic feature based technique\cite{hamidian-diab-2016-rumor, zeng2016unconfirmed}, some used conversation thread in temporal sequence\cite{zubiaga2016stance, yang2019blcu_nlp, fajcik-etal-2019-fit}, while some used it in tree-structured format\cite{wei2019modeling, pamungkas2019stance}. 
In RumorEval 2019\cite{gorrell-etal-2019-semeval}, the trend has been toward the neural approaches, with almost all the system used NN(Neural Network) based approaches except only two systems  for task B(veracity classification). The baseline system\cite{kochkina2017turing} used LSTM to model the sequential branches in the conversation  which is also ranked first in the SemEval 2017 contest. For task A, top performing systems\cite{gorrell-etal-2019-semeval} used approaches based on pre-trained models. Like in BUT-FIT\cite{fajcik-etal-2019-fit}, which ranked second, used ensemble of BERT\cite{devlin2018bert} based models. Similarly, the best performing system, BLCU-NLP\cite{yang2019blcu_nlp} and third best performing system, CLEARumor\cite{baris2019clearumor} also used the pre-trained models with BLCU-NLP\cite{yang2019blcu_nlp} used OpenAI GPT\cite{radford2018improving} and ClEARumor\cite{baris2019clearumor} used ELMo\cite{peters2018deep}. Difference lies in all system in their input structure, while BLCU-NLP\cite{yang2019blcu_nlp} uses the inference chain i.e. conversation from source post to replies (direct or indirect) arranged in time sequence. Also, linguistic feature has been used along with each post content and trained jointly for both task A and B.
For task B, the best performing system (eventAI)\cite{li-etal-2019-eventai} approached the problem using an ensemble of classifiers (SVM, LR, RF), including NN with three connected layers. Besides, other features it uses post representation obtained using LSTM with attention. The second ranked system\cite{gorrell-etal-2019-semeval} also uses the similar ensemble  with sophisticated features and feature selection using RF.
There are some previous studies which support the fact that it is necessary to solve the task A as a first step before veracity 
identification\cite{wei2019modeling, glenski-etal-2018-identifying, zubiaga2016stance, qazvinian-etal-2011-rumor, zhang2015automatic, 10.1145/1964858.1964869, procter2013reading, 10.1145/2806416.2806651}. 
\section{Problem Definition}
\label{probdef}
Consider a conversation thread $C$ originating with a source post $t_1$ followed by a number of reply posts $\{t_{2}, t_{3},....., t_{|C|}\}$ that replies $t_1$ directly or indirectly, and each post $t_i$ (i $\in$ [1, |C|] ) has a particular stance category. This paper focuses on two tasks: a) Rumor Stance Classification, aiming to determine the stance of each post $t_i$ in $C$, which belongs to \{Supporting, Denying, Querying, Commenting\}, \& b) Rumor Veracity Prediction, with the aim of identifying the veracity of the rumor, belonging to \{True, False, Unverified\}.

\section{Proposed Method}
\label{propm}
We proposed the multi-task learning framework for jointly predicting rumor stance and veracity. The joint architecture of our system is illustrated in Figure \ref{fig:arch} that is composed of two components. The first component is to classify each post in the conversation thread into four different stance labels \textit{\{support, comment, query, deny\}}. It models each post in a multi-turn conversation thread with pre-trained Longformer using sliding-window based self-attention\cite{beltagy2020longformer}. Analogous to CNNs\cite{wu2019pay}, it has multiple stacked layers resulting in a large receptive field. It outputs each of the post representation at the corresponding [CLS] token which is then concatenated with feature representations (obtained after passing the NLP Features (see Table. \ref{tab:feat}) through a linear layer). This feature augmented post representation is then input to Sentence Encoder(see Section\ref{sentenc}) to classify each post into the four different stance labels \textit{\{support, comment, deny, query\}}. The second component is to classify the rumor’s veracity into three labels \textit{\{True, False, Unverified\}} by taking the 1-dimensional mean-pooling of post representations at the output of Sentence Encoder.

\subsection{Pre-processing}
We have normalized the text to make it suitable for feeding to the Longformer\cite{beltagy2020longformer}. Same pre-processor as in\cite{fajcik-etal-2019-fit} is used to normalize the text except the tokenizer. Instead, we have used the tokenizer from Hugging Face PyTorch re-implementation of Longformer\footnote{https://huggingface.co/transformers/model\_doc/longformer.html}. Specifically, we have used tweet-processor\footnote{https://github.com/s/preprocessor} to identify URLs and mentions and replace them with special tokens \$URL\$ and \$mention\$, and spcay\footnote{https://spacy.io/} has been used to split each post into sentences and add the [EOS] token to specify the termination of sentence. 

\subsection{Feature Extraction}
\label{feat}
We have identified a novel combination of features that are highly effective to provide indicative clues for rumor stance and veracity when learning the fine-tuned representation on top of the pre-trained model. We have introduced two new features in addition to previously available features. The first one is the use of Paragram embeddings\cite{wieting2015towards} to get each of the post representations, and hence-forth similarity with previous, source, and other posts joined in temporal sequence. The second one is the speech-act categories\cite{wierzbicka1987english} containing the collection of 229 verbs divided into 37 categories. (see Table \ref{tab:feat})\\
Different from previous approaches in RumorEval 2019\cite{gorrell-etal-2019-semeval} we have extracted various psycho-linguistic features like emotion features (from Emolex\cite{mohammad2013crowdsourcing, ekman1992argument}, EmoSenticnet\cite{poria2013enhanced, ekman1992argument} and LIWC\cite{pennebakerlinguistic}), affective features (ANEW\cite{osgood1957measurement} and AFINN\cite{nielsen2011afinn}). Different from\cite{yang2019blcu_nlp} which extracted the features (false synonym, false antonym, number of question words, presence of rumor words and words indicating absence of assertion) for each post in the conversation, but in our case, to highlight the difference between the content of source post and any thread post in conversation, the features extracted for each post is the concatenation of features from source post and thread post. In case, to highlight the linguistic structure an additional 37 sized vector indicating presence of each of the pos tags\footnote{36 from https://www.ling.upenn.edu/courses/Fall\_2003/ling001/penn\_treebank\_pos.html \& ’X’ from https://spacy.io/api/annotation\#pos-tagging}. All the features has been extracted on raw text as it is in the dataset provided by RumorEval 2019\cite{gorrell-etal-2019-semeval} except only the feature named \enquote{Conversational Features} that has extracted from the processed text. The features extracted has been passed through a linear layer of size $d_2$, hence $f_i \in R^{d_2} \textrm{   } \forall \textrm{    } i \in \textrm{    }  [1, |C|]$.

\begin{table*}[]
\resizebox{\textwidth}{!}{\begin{tabular}{|l|l|l|}
\hline
\multicolumn{1}{|c|}{\textbf{Feature Name}} & \multicolumn{1}{c|}{\textbf{Description}}                                                                                                                                                                                                                                                                                                                                                                                                                                                                                                    & \multicolumn{1}{c|}{\textbf{Feature Count}} \\ \hline
Structural Features                         & \begin{tabular}[c]{@{}l@{}}The structural features for each post extracted are: a) Feature vector indicating presence of 37 POS tags\setfootnotemark\label{first}.\\ b) Presence of exclamation mark, negative words, media content like pic, video etc, URL, period,  question mark, \\ hashtag. c) Count of chars, words and ratio of capital letters. d) Whether post is the source post. e) Count of exclama-\\ tion marks, question marks, and periods.\end{tabular}                                                                     & 51                                          \\ \hline
Content Features                            & \begin{tabular}[c]{@{}l@{}}The features extracted to capture the content of each post are: a) 2 features indicating presence of cue\cite{bahuleyan-vechtomova2017semeval} and swear \\ words\setfootnotemark\label{second} b) Count of false synonyms, antonyms and question word for each post concatenated with that \\ of source post thus 6 features. c) Presence of rumor words like 'gossip', 'hoax' etc. for both source and thread post.\\ d) Presence of words indication unsure about assertion like 'unconfirmed','unknown' for both source and thread post.\end{tabular} & 12                                          \\ \hline
Conversational Features                     & \begin{tabular}[c]{@{}l@{}}Peculiar to tree-structured conversation we have extracted various features a) Paragram Embedding\cite{wieting2015towards} \\ representation for each post in the conversation.(300) b) Similarity between thread post and source post \\ c) Similarity between thread post and previous post d) Similarity with concatenated posts (other than source \\ and previous) in time order e) Normalized Depth of post in the tree structured conversation.\end{tabular}                                                & 305                                         \\ \hline
Affective Features                          & \begin{tabular}[c]{@{}l@{}}There are seven features extracted from three resources a) Three from DAL\cite{whissell2009using} b) Three from ANEW\cite{osgood1957measurement} \\ and c) One is the sentiment score from AFINN\cite{nielsen2011afinn}.\end{tabular}                                                                                                                                                                                                                                                                                 & 7                                           \\ \hline
Emotion Features                            & \begin{tabular}[c]{@{}l@{}}Extracted 8 primary emotion based on Plutchik model\cite{plutchik2001nature, mohammad2013crowdsourcing}, extra 2 signals is also extracted indicating \\ positive and negative emotion. Additionally, 6 basic emotion categories from EmoSenticNet\cite{poria2013enhanced, ekman1992argument}.\end{tabular}                                                                                                                                                                                                                                                      & 16                                          \\ \hline
LIWC                                        & \begin{tabular}[c]{@{}l@{}}There are 11 categories of LIWC\cite{pennebakerlinguistic} are further grouped according to four stance categories as \\ a) agree-accept(Support) : Assent, Certain, Effect; b) Reject(Deny): Negate, Inhib; \\ c) Info-request(Query): You, Cause; d) Opinion(Comment): Future, Sad, Insight, Cogmech. Additionally, \\ two categories indicating positive and negative sentiment is also extracted.\end{tabular}                                                                                                   & 13                                          \\ \hline
Speech-Act Features                         & \begin{tabular}[c]{@{}l@{}}Certain verbs like ask, demand, promise report etc. that categorize the different speech acts, also able\\  to indicate the stance category which in turn indicate the rumor veracity. Extracted 37 categories\cite{wierzbicka1987english} which \\ compiled a total of 229 verbs across different categories.\end{tabular}                                                                                                                                                                                                                   & 37                                          \\ \hline
Total Features                              &                                                                                                                                                                                                                                                                                                                                                                                                                                                                                                                                              & 441                                         \\ \hline
\end{tabular}}
\caption{NLP Features}
\label{tab:feat}
\afterpage{\footnotetext[\getrefnumber{first}]{36 from https://www.ling.upenn.edu/courses/Fall\_2003/ling001/penn\_treebank\_pos.html and 'X' from https://spacy.io/api/annotation\#pos-tagging}
             \footnotetext[\getrefnumber{second}]{https://www.cs.cmu.edu/ biglou/resources/bad-words.txt}}
\vspace{-6mm}
\end{table*}

\subsection{Encoding each utterance in conversation thread}
As mentioned in Section \ref{intro}, the nearest neighbors of a post provide a more informative signal for the stance of a post. Based on the above information we proposed to model the structural and temporal property to learn the stance feature representation of each post in the conversation thread. For that, it is required to give a complete conversation as input. Since the conversations can be arbitrarily large (Reddit conversations are usually larger than those of twitter (see Section \ref{intro}) ) we decided to use the Longformer\cite{beltagy2020longformer} which is the current state-of-art for long contexts datasets like Wikihop\cite{welbl2018constructing}, HotPotQA\cite{yang2018hotpotqa}, TriviaQA\cite{joshi2017triviaqa}, etc. 

To use Longformer\cite{beltagy2020longformer} for encoding each post in the conversation thread requires it to output the representation of each sentence/post in the conversation. However, since Longformer (based on Roberta\cite{liu2019roberta}) is trained as a
masked-language model, the output vectors are grounded to tokens instead of sentences (or post-level in this case).  Therefore, we modify the input sequence of Longformer\cite{beltagy2020longformer} to make it possible for extracting post representations.

\subsubsection*{\textbf{Encoding Multiple Sentences}} As illustrated in
Figure \ref{fig:arch}, we insert a [CLS] token before each sentence and a [SEP] token after each sentence. In vanilla Longformer (which is trained from Roberta\cite{liu2019roberta} checkpoint), the [CLS] token is used to aggregate features from one sentence or a pair of sentences using global attention\cite{beltagy2020longformer}. We modify the model by using multiple [CLS] tokens to get a sentence vector (post representation) using local attention based on the sliding window. After obtaining the sentence vectors from Longformer\cite{beltagy2020longformer} each of the vector $t_i \in R^{d_1}$ has been concatenated with their corresponding feature representation $f_i \in R^{d_2}$ given as 
\begin{equation}
    k_i =  t_i \oplus f_i \textrm{          } \in R^{d_1 + d_2} \textrm{                     }\forall i \in [1, |C|]
\end{equation}
\subsection{Sentence Encoder}
\label{sentenc}
The feature augmented vectors $k_i \in R^{d_1 + d_2}$ are then fed to the encoder-specific layers stacked on top of Longformer\cite{beltagy2020longformer}. These encoder layers are jointly fine-tuned with Longformer\cite{beltagy2020longformer} to learn the temporal evolution of stance in the conversation (see Figure \ref{fig:arch}). The various encoders we have experimented with are described as follows:
\subsubsection*{\textbf{Identity Encoder}}
We have used this encoder for the sake of comparison with other encoders and to compare the robustness of the representation we get at the output of Longformer\cite{beltagy2020longformer} versus other encoders. Encoded representation, in this case, is just $k_i$ only given as:
\begin{equation}
    h_i = k_i \in R^{m} \textrm{   } \forall \textrm{    } i \in [1,2...|C|]
\end{equation}
In this case, $m = d_1 + d_2$.
\subsubsection*{\textbf{Inter-Sentence Transformer}}
Inter-Sentence Transformer applies additional Transformer\cite{vaswani2017attention} layers only on feature augmented post representation i.e. $k_i$ to extract the relation between post and their corresponding stances. The encoded representation for a layer $l$ is given as follows:
\begin{align}
    {\Tilde{g}_i}^l = LN ( {g_i}^{l-1} + MHATT ( {g_i}^{l-1}))
    \\
{g_i}^{l} = LN({\Tilde{g}_i}^l + FFN({\Tilde{g}_i}^l ))
\end{align}
where ${g_i}^0 = k_i \oplus PosEMB(K)$, where $K = [k_1, k_2, ..... k_{|C|}]$. Here, the symbols like PosEMB, LN, FFN, MHATT are the function to add Positional Embeddings,  Layer Normalization, Feed-Forward Network, Multi-Head Attention respectively. These function implementation is taken as it is described in Transformer\cite{vaswani2017attention}. Let there be $L$ number of layers, final output is given as follows:
\begin{equation}
    h_i = {g_i}^L \textrm{      } \in R^m \textrm{    } \forall i \in [1,2...|C|]
\end{equation}
In experiments, we implemented Transformers with $L = 1, 2, 3$ and found Transformer\cite{vaswani2017attention} with 2 layers performs the best.
\subsubsection*{\textbf{Recurrent Neural Network(RNN)}} 
Although, the pre-trained transformers achieved state-of-art on several tasks. But, RNN stacked over transformer seem to achieve better results\cite{chen2018best}. We have used LSTM\cite{hochreiter1997long} in this case to learn the task specific representation. Similar to\cite{ba2016layer} we have also applied Layer normalization per-gate to each LSTM cell. 
\begin{align}
\begin{pmatrix}
F_i\\ 
I_i\\ 
O_i\\ 
G_i
\end{pmatrix} = {LN}_h(W_h r_{i-1}) + {LN}_x(W_x k_i)
\\
C_i = \sigma(F_i) \odot C_{i-1}
+ \sigma(k_i) \odot tanh(G_{i-1})
\\
r_i = \sigma(O_t) \odot tanh({LN}_c(C_t))
\end{align}
where $F_i, I_i, O_i$ are forget gates, input gates, output gates; $G_i$ is the hidden vector and $C_i$
is the memory vector; $r_i$ is the output vector; $LN_h, LN_x, LN_c$ are the layer normalization operations at output, input and memory vector respectively;  Bias terms are not shown. The final output vector after linear layer is given as:
\begin{equation}
    h_i = r_i \textrm{      } \in R^m  \textrm{     } \forall i \in [1,2...|C|]
\end{equation}
\subsection{Rumor Stance and Veracity Classification}
After getting the fine-tuned post  representation $h_i$ from the sentence encoder, it has been used to classify the stance of each post and rumor veracity each has been described as follows:
\subsubsection*{\textbf{Stance Classification}} 
For stance classification each of the fine-tuned representation $h_i \textrm{      } \in R^{m} \textrm{      } \forall \textrm{   } i \in \textrm{   } [1,2,....|C|]$ is to be classified among four labels \textit{\{support, comment, deny, query\}} numbered as $[0,1,2,3]$. For each post $t_i$ in the conversation $C$, we apply softmax to obtain its predicted stance distribution:
\begin{equation}
    \hat{y}^{st}_i = softmax(W_{st} h_i  + b_{st})
\end{equation}
where $W_{st} \in R^{4 \times (m)}$ and $b_{st} \in R^4$ are weight matrix and bias respectively. The loss function of $C$ for stance classification is computed by cross-entropy criterion:
\begin{equation}
    \mathcal{L}_{stance} = \frac{1}{|C|}\Sigma_{i=1}^{|C|}-{({y_i}^{st})}^T \log \hat{y}_i^{st}
\end{equation}
where gold label $y_i^{st}$ is the one-hot vector denoting the stance label for the post $t_i$. For batched training the cross-entropy loss is the average cross-entropy over the number of examples in a batch.
\subsubsection*{\textbf{Veracity Classification}} 
The fine-tuned post representation vectors $\{h_1, h_2, h_3,.....h_{|C|}\}$ are the output sequence that represents the temporal feature. We then transform this temporal sequence to a vector $v$ by a 1-dimensional global mean-pooling to capture the stance evolution. The mean-pooled representation 
$\hat{\boldsymbol{h}}$ is then used for veracity classification after passing through linear layer and softmax normalization. For veracity classification, there are three labels \textit{\{True, False, Unverified\}} numbered as $[0, 1, 2]$. The predicted distribution over veracity labels is given as:
\begin{align}
    \hat{\boldsymbol{h}} = \textrm{mean-pooling}(h_1, h_2, ....h_{|C|}),
    \\
    \hat{y}_i^{ver}  = softmax( W_{ver} \boldsymbol{\hat{h}} + b_{ver} )
\end{align}
where $W_{ver} \in R^{3 \times (m)}$ and $b_{ver} \in R^3$ are weight matrix and bias respectively. The cross-entropy loss function of $C$ for veracity classification is given as:
\begin{equation}
    \mathcal{L}_{veracity} = - y_i^{ver} \log \hat{y}_i^{ver}
\end{equation}
where the gold label $y_i^{ver}$ is the one-hot vector denote the veracity label for the rumor started with post $t_0$.
\subsection{Jointly Learning Two Tasks}
As mentioned in Section \ref{intro}, the stance evolution indicates the rumor veracity so we should leverage the interrelation between the two tasks i.e. stance classification and subsequent task which is veracity classification. We have trained these two tasks jointly by adding the loss function for each task with a trade-off parameter $\lambda$ and optimize them jointly. Specifically, the joint loss $\mathcal{L}$ is given as: 
\begin{equation}
    \mathcal{L} = \mathcal{L}_{veracity} + \lambda   \mathcal{L}_{stance}
\end{equation}
\subsection{Ensembling}
Our overall architecture of model consists of Longformer\cite{beltagy2020longformer} as a base model on top of that various sentence encoder has been put to learn the varying features for stance classification and on top of that, we leverage the stance evolution for veracity classification. So, we have trained and finally saved the 50 best models by varying the learning rates and encoders. We have used the Top-$N_s$ fusion strategy as described in\cite{fajcik-etal-2019-fit} in order to increase the F1 measure and reduce overfitting. This procedure iteratively selects 1 model after random shuffling the pool of models and adds it to the ensemble, if it increases the ensemble's F1 by averaging the output probabilities, effectively approximating the  Bayesian model averaging. Specifically, in Top-$N_s$ strategy we take the average of pre-softmax scores instead of output probabilities.

\section{Experiments}
\label{exp}
In this section, we first evaluate the performance of rumor stance classification and then veracity
prediction (Section \ref{evalstrat}). We then give a detailed analysis of our proposed method (Section \ref{resdis} and \ref{case}).
\begin{table}[]
\begin{tabular}{|l|l|l|l|l|l|}
\hline
              & \textbf{Support} & \textbf{Deny} & \textbf{Query} & \textbf{Comment} & \textbf{Total} \\ \hline
Twitter Train & 1004             & 415           & 464            & 3685             & 5568           \\
Reddit Train  & 23               & 45            & 51             & 1015             & 1134           \\ \hline
Total Train   & 1027             & 460           & 515            & 4700             & 6702           \\ \hline
Twitter Test  & 141              & 92            & 62             & 771              & 1066           \\
Reddit Test   & 16               & 54            & 31             & 705              & 806            \\ \hline
Total Test    & 157              & 146           & 93             & 1476             & 1872           \\ \hline
Total Task A  & 1184             & 606           & 608            & 6176             & 8574           \\ \hline
\end{tabular}
\caption{Task A Corpus}
\label{tab:taskadata}
\vspace{-6mm}
\end{table}
\begin{table}[]
\begin{tabular}{|l|l|l|l|l|}
\hline
              & \textbf{True} & \textbf{False} & \textbf{Unverified} & \textbf{Total} \\ \hline
Twitter Train & 145           & 74             & 106                 & 325            \\
Reddit Train  & 9             & 24             & 7                   & 40             \\ \hline
Total Train   & 154           & 98             & 113                 & 365            \\ \hline
Twitter Test  & 22            & 30             & 4                   & 56             \\
Reddit Test   & 9             & 10             & 6                   & 25             \\ \hline
Total Test    & 31            & 40             & 10                  & 81             \\ \hline
Total Task B  & 185           & 138            & 123                 & 446            \\ \hline
\end{tabular}
\caption{Task B Corpus}
\label{tab:taskbdata}
\vspace{-8mm}
\end{table}

\subsection{Data \& Evaluation Metric}
We have used the data released at RumorEval 2019 for both subtask A(stance classification) and B(veracity classification)\cite{gorrell-etal-2019-semeval}. For each task the distribution of train and test is shown in Tables \ref{tab:taskadata} and \ref{tab:taskbdata}. We have used the same evaluation metric as in RumorEval 2019\cite{gorrell-etal-2019-semeval} i.e. Macro-averaged F1 score and RMSE (For task A only).
\begin{table*}[ht]
\begin{minipage}{0.49\linewidth}
            \centering
\resizebox{\textwidth}{!}{%
\begin{tabular}{|l|l|l|l|}
\hline
\multicolumn{1}{|c|}{\textbf{Rank}} & \multicolumn{1}{c|}{\textbf{System}}                                                       & \multicolumn{1}{c|}{\textbf{MacroF}} & \multicolumn{1}{c|}{\textbf{RMSE}} \\ \hline
1                                   & eventAI                                                                                    & 0.5765                               & 0.6078                             \\ \hline
2                                   & WeST (CLEARumor)                                                                           & 0.2856                               & 0.7642                             \\ \hline
3                                   & GWU NLP LAB                                                                                & 0.2620                               & 0.8012                             \\ \hline
4                                   & BLCU NLP                                                                                   & 0.2525                               & 0.8179                             \\ \hline
5                                   & shaheyu                                                                                    & 0.2284                               & 0.8081                             \\ \hline
\multicolumn{4}{|c|}{\textbf{Our Models}}                                                                                                                                                                    \\ \hline
A                                   & Longformer + Identity Encoder                                                                                &         
0.3795                        &     0.7240                               \\ \hline
B                                   & Longformer + Transformer                                                                   &    
0.3363                            &     0.7212                               \\ \hline
C                                   & Longformer + Bi-LSTM                                                                       &   
0.4004                                   &  0.7394                                  \\ \hline
D                                   & \begin{tabular}[c]{@{}l@{}}Longformer + Identity Encoder\\ + NLP Features\end{tabular}                                                                  &     
0.4962                                 &      0.6577                              \\ \hline
E                                   & \begin{tabular}[c]{@{}l@{}}Longformer + Transformer\\ + NLP Features\end{tabular}          & 
0.5327
                                     &      0.6299                              \\ \hline
F                                   & \begin{tabular}[c]{@{}l@{}}Longformer + Bi-LSTM \\ + NLP Features\end{tabular}             &            
0.5275
                          &     0.6291                               \\ \hline
                                    & \begin{tabular}[c]{@{}l@{}}Our proposed method - Top $N_s$\\ using (D + E + F)\end{tabular} & 0.5868                                 &   0.6056
                                  \\ \hline
\end{tabular}
}
\caption{Test results for Task B}
\label{tab:taskb}
\end{minipage}
\hfillx
\begin{minipage}{0.42\textwidth}
            \centering
\resizebox{\textwidth}{!}{%
\begin{tabular}{|l|l|l|}
\hline
\multicolumn{1}{|c|}{\textbf{Rank}} & \multicolumn{1}{c|}{\textbf{System}}                                                        & \multicolumn{1}{c|}{\textbf{MacroF}} \\ \hline
1                                   & BLCU NLP                                                                                    & 0.6187                               \\ \hline
2                                   & BUT-FIT                                                                                     & 0.6167                               \\ \hline
3                                   & eventAI                                                                                     & 0.5776                               \\ \hline
4                                   & UPV-28-UNITO                                                                                & 0.4895                               \\ \hline
5                                   & HLT(HITSZ)                                                                                  & 0.4792                               \\ \hline
\multicolumn{3}{|c|}{\textbf{Our Models}}                                                                                                                                \\ \hline
A                                   & Longformer + Identity Encoder                                                                                 &   0.5782                              \\ \hline
B                                   & Longformer + Transformer                                                                    &   0.5807                                 \\ \hline
C                                   & Longformer + BiLSTM                                                                         &   0.5886                                   \\ \hline
D                                   & \begin{tabular}[c]{@{}l@{}}Longformer + Identity Encoder\\ + NLP Features\end{tabular}                                                                   &    0.6371                                  \\ \hline
E                                   & \begin{tabular}[c]{@{}l@{}}Longformer + Transformer\\ + NLP Features\end{tabular}           & 0.6389                                      \\ \hline
F                                   & \begin{tabular}[c]{@{}l@{}}Longformer + BiLSTM\\ + NLP Features\end{tabular}                &            0.6487                          \\ \hline
                                    & \begin{tabular}[c]{@{}l@{}}Our Proposed Method - Top $N_s$ \\ using (D + E + F)\end{tabular} &    0.6720                                 \\ \hline
\end{tabular}
}
\caption{Test results for Task A}
\label{tab:taska}
\end{minipage}
\\
\begin{minipage}{0.7\linewidth}
\resizebox{\linewidth}{!}{%
\begin{tabular}{|l|l|l|l|l|l|l|}
\hline
\multicolumn{7}{|c|}{\textbf{Stance Classification - Class-wise F1 Scores}}                                                                                                                                                                                                                                                                      \\ \hline
\multicolumn{2}{|l|}{}                                                                                                  & \multicolumn{1}{c|}{\textbf{$F1_{S}$}} & \multicolumn{1}{c|}{\textbf{$F1_{D}$}} & \multicolumn{1}{c|}{\textbf{$F1_{Q}$}} & \multicolumn{1}{c|}{\textbf{$F1_{C}$}} & \multicolumn{1}{c|}{\textbf{$macroF_{test}$}} \\ \hline
\multicolumn{2}{|l|}{\begin{tabular}[c]{@{}l@{}}Our Proposed method \\ ( Top-$N_{s}$ using (D + E + F) )\end{tabular}} &         0.5158                                   & 0.9256                                        &        0.5890                                  &                  0.6576                       & 0.6720                                               \\ \hline
\multicolumn{7}{|c|}{\textbf{Veracity Classification - Class wise F1 Scores}}                                                                                                                                                                                                                                                                    \\ \hline
\multicolumn{2}{|l|}{}                                                                                                  & \textbf{$F1_{True}$}                   & \textbf{$F1_{False}$}                  & \multicolumn{2}{l|}{\textbf{$F1_{Unverified}$}}                                  & \textbf{$macroF_{test}$}                      \\ \hline
\multicolumn{2}{|l|}{\begin{tabular}[c]{@{}l@{}}Our Proposed method \\ ( Top-$N_{s}$ using (D + E + F) )\end{tabular}} &                    0.4651                     &            0.7238                             & \multicolumn{2}{l|}{0.5715}                                                             &              0.5868                                 \\ \hline
\end{tabular}
}
\caption{Individual F1 score for each class}
\label{tab:indiv}
\end{minipage}
\vspace{-6mm}
\end{table*}
\subsection{Implementation Details}
We implemented our models in Pytorch using Hugging Face implementation of Longformer\footnote{https://huggingface.co/transformers/model\_doc/longformer.html} with pre-trained parameters loaded from \enquote{longformer-base-4096} having 12 transformer layers, hidden unit size of d = 768, 12 attention heads, vocab size = 50265, max length = 4096. Longformer, sentence encoder, and task related classification layers are jointly fine-tuned. For conversation  which does not fit into the max length of 4096, we have created multiple examples using the sliding window at post level (adding one post at a time and remove from source end). Since the sentence encoder and classification layers have to be trained from scratch, while Longformer is already pre-trained this may cause instability for example pre-trained one may overfit the data while sentence encoder underfits. Therefore, we have used two Adam optimizers with $\beta_{1} = 0.9$ and $\beta_{2} = 0.999$ for pre-trained (P) and other components (OC) respectively, each with different warmup-steps and learning rates:
\begin{align}
    lr_P = \Tilde{lr}_P \cdot min(step^{-0.5}, step \cdot warmup^{-1.5}_{P} )
    \\
lr_{OC} = \Tilde{lr}_{OC} \cdot min(step^{-0.5} , step \cdot warmup^{-1.5}_{OD}) 
\end{align}
The size of the Linear Layer used to obtain feature representation is taken as 128. In the case of Transformer as sentence encoder, output hidden size is same as the input one(in our case 768+128) and the number of layers = 2, while in case of Bi-LSTM output size is 512 where 256-dimensional vector is obtained from each direction. Also, for joint training $\lambda = 0.7$ gives the best results.
\begin{figure*}[t]
    \centering
    \includegraphics[width=0.7\textwidth]{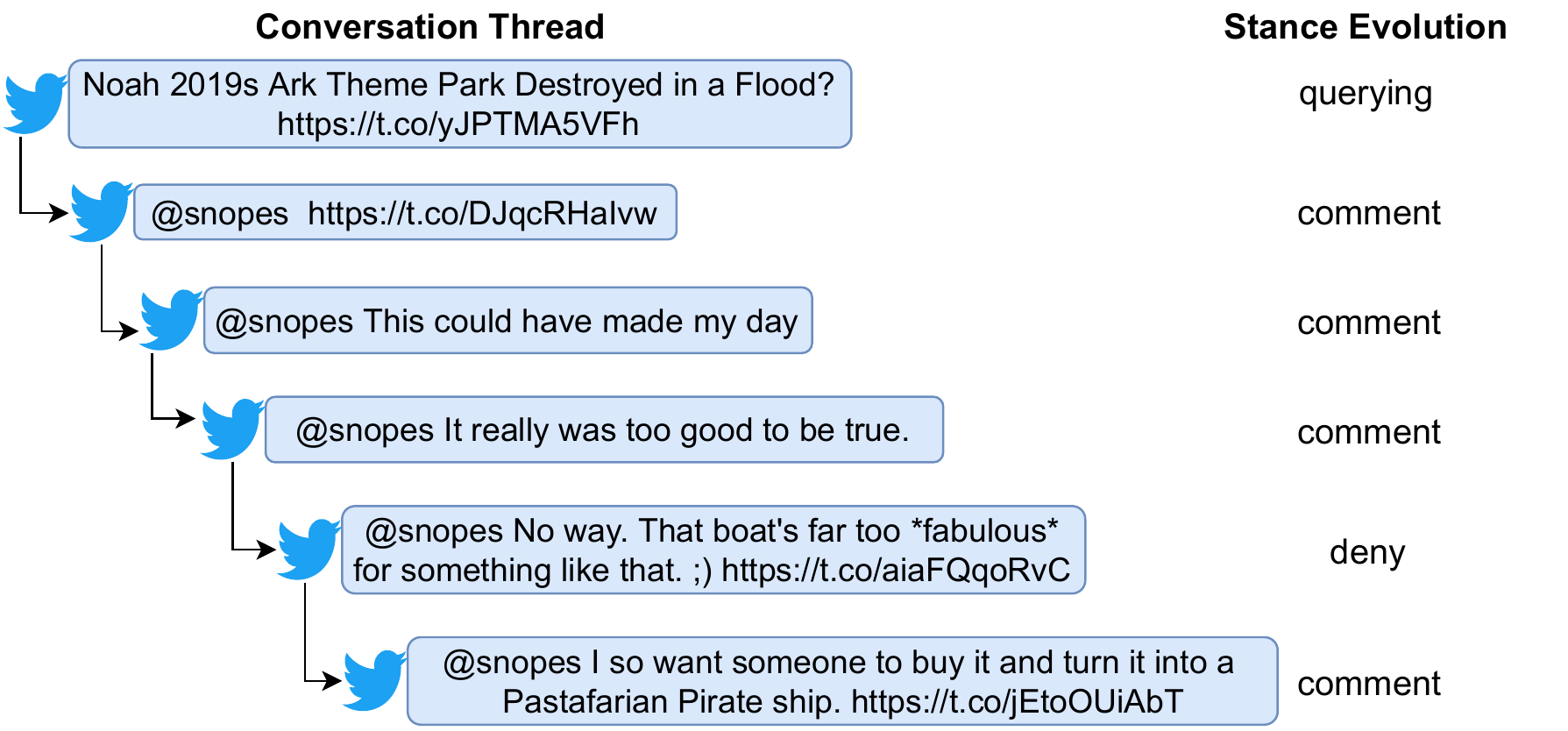}
    \caption{Case Study: a \textit{false} rumor}
    \label{fig:case}
    \vspace{-5mm}
\end{figure*}
\subsection{Evaluation Strategy}\label{evalstrat}
In this section we have evaluated our proposed approach against the Top-5 systems submitted at RumorEval 2019\cite{gorrell-etal-2019-semeval} for Task A and Task B. We have conducted the separate experiments, to properly investigate the performance of a) each of the classifiers used in the ensemble. b) impact of NLP features on each of those classifiers and finally, c) the performance of our proposed system. In Tables \ref{tab:taska}, \ref{tab:taskb} and \ref{tab:indiv}, models named as \textbf{Longformer + Identity Encoder}, \textbf{Longformer + Transformer} and, \textbf{Longformer + BiLSTM} are corresponding models with different types of sentence encoders (see Section \ref{sentenc}) over the base model Longformer (see Figure \ref{fig:arch}) without concatenating NLP features (see Section \ref{feat}) at the fine-tuned post representation as shown in Figure \ref{fig:arch}.
Similarly, \textbf{Longformer + Identity Encoder + NLP Features}, \textbf{Longformer + Transformer + NLP Features}, \textbf{Longformer + BiLSTM + NLP Features} are corresponding models with NLP features. Specifically, for the model named \textbf{Longformer + Identity Encoder} is only a base model (Longformer) without any encoder to judge the performance boost as compared to when we use BiLSTM or Transformer encoder. Here, NLP Features have been applied to Linear Layer(of size 128), output of which has been concatenated to fine-tuned representation obtained at the output of sentence encoder (see Section \ref{sentenc}). In addition we have reported the class wise results of best model for both tasks as shown in Table \ref{tab:indiv}. \textbf{Top-$N_s$ (D +  E + F)} is the ensembling strategy based on Model Averaging\cite{fajcik-etal-2019-fit} of the selected models from the pool of models saved after varying the learning rates and encoder. Each model is chosen randomly and if it increases the ensemble's F1 then it has been added to the ensemble. To create the pool of models, we consider only models trained with NLP features since they have better performance, i.e. D, E \& F. Finally, \textbf{Our proposed method} represents the model averaging of models based on three architectures(D, E \& F) with NLP Features trained with varying parameters (encoder and  learning rate). 

\subsection{Results and Discussion}\label{resdis}
In this paper, we have evaluated our models using the same guidelines as in RumorEval 2019 contest paper\cite{gorrell-etal-2019-semeval}. Specifically, they have used macro-averaged F1 to evaluate the performance on both Tasks A and B. Additionally, they have used the RMSE score for Task B to judge the confidence scores. We followed the same guidelines as provided by the baseline system\cite{kochkina2017turing} to calculate the score for our proposed system. Tables \ref{tab:taska}, \ref{tab:taskb}, and \ref{tab:indiv} presents the comparative experimental results for the proposed method in this paper with respect to the state-of-the-art. The Top-5 systems\cite{gorrell-etal-2019-semeval} given in Table \ref{tab:taska} and \ref{tab:taskb} are the best-performing systems as per the published results in RumorEval 2019 paper\cite{gorrell-etal-2019-semeval}. From the results, given in the Tables \ref{tab:taska}, \ref{tab:taskb} and, \ref{tab:indiv} it is clear that our proposed method shows the best performance among all the approaches. These results also state the importance of NLP Features and Sentence Encoder. We will discuss the effect of each in the following sections.

\subsubsection*{\textbf{Effect of NLP Features}}
To understand the importance of NLP Features, we conduct an ablation study: we only input the post representation to the classification layer with or without sentence encoder(ref. Table \ref{tab:taska}, \ref{tab:taskb}). The results state that the sentence encoder only models the temporal variation of post representation but not able to capture the cause for the particular category for stance or veracity.

\subsubsection*{\textbf{Effect of Sentence Encoder}}
We have studied the impact of using the sentence encoder with or without NLP Features using the model named as \textbf{Longformer + Identity Encoder} and \textbf{Longformer + Identity Encoder + NLP Features}. In both cases, the performance in terms of F1 is less than as compared to when we use the encoder (either BiLSTM or Transformer). The results state that the sentence encoders helps to learn the stance evolution to determine the proper category of veracity and neighbouring posts helps in determining the stance category.

\subsection{Case Study}
\label{case}
An example of the \enquote{\textit{false}} rumor identified by our model is illustrated in Figure \ref{fig:case}. It illustrates the conversation thread starting from source post followed by subsequent replies in the temporal order. As seen, the stance evolution contains a sequence \enquote{\textit{query} \textrightarrow \textit{comment} \textrightarrow \textit{comment} \textrightarrow \textit{comment} \textrightarrow \textit{deny}}. Since, the source post does not indicate of why this should be a false rumor, our model captures the stance evolution using fine-tuned representations obtained at the output of sentence encoder (see Section \ref{sentenc}), and hence accumulated information using average pooling correctly identifies the rumor veracity.
\section{Conclusion and Future Work}\label{conc}
In this paper, we have briefly described the multi-task approach for joint prediction of rumor stance and veracity for data obtained from various social media platforms (in our case Twitter and Reddit). We have presented an ensemble of deep learning models having the same architecture, but varying the parameters. Our approach outperforms all the previous approaches by a sufficient margin and able to generalize across different social media.
In future, we can extend our model for multilingual setting\cite{wen-etal-2018-cross} (IberEval is the counterpart of RumorEval 2019 for other languages\cite{gorrell-etal-2019-semeval}). Moreover, we can leverage more sophisticated resources like a pre-trained model which was trained specifically to handle data from different social media platforms. Further, we can explore other methods like diffusion process of rumors\cite{vosoughi2018spread} to make informed changes to model architecture. 
\newpage
\bibliographystyle{ACM-Reference-Format}
\bibliography{codscomad-23}
\end{document}